\documentclass{article}
\usepackage[utf8]{inputenc}
\usepackage{natbib}
\usepackage{gb4e}
\usepackage{paralist}
\usepackage{hyperref}
\usepackage{authblk}

\title{\emph{Rat big, cat eaten!}\\Ideas for a useful deep-agent protolanguage}
\author[1,2]{Marco Baroni}
\affil[1]{Facebook AI Research}
\affil[2]{Catalan Institution for Research and Advanced Studies}

\begin{document}

\maketitle

\begin{abstract}
  Deep-agent communities developing their own language-like
  communication protocol are a hot (or at least warm) topic in
  AI. Such agents could be very useful in machine-machine and
  human-machine interaction scenarios long before they have evolved a
  protocol as complex as human language. Here, I propose a small
  set of priorities we should focus on, if we want to get as fast as
  possible to a stage where deep agents speak a useful
  \emph{protolanguage}.
\end{abstract}

\section{Introduction}

Deep-agent emergent language research rests on the hypothesis that, if we
want deep networks to develop language-like communication skills, we
cannot just train them to reproduce statistical regularities in static
linguistic corpora (as is done in language modeling, e.g.,
\cite{Radford:etal:2019}), but we should plunge them into interactive
scenarios, letting them develop a code to cooperatively solve
their tasks (e.g.,
\cite{Foerster:etal:2016,Lazaridou:etal:2017,Cao:etal:2018,Havrylov:Titov:2017,Kottur:etal:2017,Evtimova:etal:2018,Mordatch:Abbeel:2018,Chaabouni:etal:2019,Li:Bowling:2019,Lowe:etal:2019},
among many others). I assume here that this is a worthy research
program, and that anybody reading this has a minimal degree of interest
in it. \citet{Lazaridou:Baroni:2020} provide a recent overview of the
area.

% Now, if our long-term goal is indeed to develop agents that can
% effectively cooperate with each other and with humans through
% language, what are the first steps we should take in our research?

Both on the phylogenetic and on the ontogenetic scale, human language
does not appear all at once in fully-formed garb. Most linguists agree
that, as a species, we went through a protolanguage stage involving a
small set of simple constructions
(\cite{Bickerton:2014,Brentari:GoldinMeadow:2017,Hurford:2014,Jackendoff:Wittenberg:2014}). Children
definitely pass through fairly systematic protolanguage phases, such
as the ``two-word'' stage
(\cite{Bloom:1970,Berk:LilloMartin:2012}). Still,
children and, presumably, our ancestors manage(d) to get a lot done
with their protolanguages.\footnote{I am not implying
   that there is any direct relation between the hypothetical
  original human protolanguage and simplified language usage by
  children. There probably isn't (\cite{Brochhagen:2018}).} Similarly, it is productive to ask:
how should a \emph{useful} protolanguage for deep agents look like?
And, consequently, what should be our priorities in terms of the
linguistic abilities we want the agents to develop?\footnote{The study
  of emergent communication has also other worthy goals, such as
  providing insights into the origins and universal properties of
  human communication. The priorities I suggest here are strictly from
  the point of view of my idea of a ``useful'' protolanguage, but of
  course other topics might be more important when emergent
  communication is studied in a different, less applied perspective.}

Since these are daunting tasks, rather trying to start from abstract,
top-down considerations, I will begin by laying out a prototypical
scenario in which an emergent language could be deployed, and analyze
which environmental and communication needs are important in it.

\section{A prototypical scenario and its demands}

Here is a concrete scenario that might be only a few years away. A community of
differently specialized bots (embodied, virtual, or a
mixture of the two) help us in our daily life, going through chores
such as checking what we have in the refrigerator, seeing whether the
water is boiling, finding our phone for us, etc. The bots will need to
communicate with each other, and at least occasionally with
us.

Scripting is not an option, because we are stuffing our cupboard with
fancy new cereals all the time; we buy new gadgets that then we
promptly lose under the sofa; and so on. Not only new things will
continuously pop up, but words will have to frequently be combined in
novel ways, with subtle semantic shifts that would be extremely
difficult to hard-wire: different things in the refrigerator might rot
in different ways, a dishwasher and a pot will be considered full
under rather different conditions, etc.\footnote{I assume that we want
  agents to speak a language that, like ours, is ambiguous,
  context-dependent, fuzzy, vague. Fully explicit, unambiguous,
  logically-consistent codes might be useful in small, well-defined
  domains, but they are too rigid for open-ended everyday life
  scenarios, witness the failure of good-old-AI. Moreover, if we want
  the bots to talk with us, we certainly do not want them to speak in
  logics.}

This is of course just an example of a scenario in which to deploy
talking bots, but I think its demands are fully representative of
those of many other useful interactive use cases, from purely virtual
chatbots to colonies of fully embodied worker robots.

The proposed scenario requires formidable progress in
robotics (how do you bend to look for the lost gadget under the
sofa?), computer vision (how do you visually recognize a new cereal?
how do you transfer rotting cues from chicken to tomatoes?) and other
fields. My focus here is exclusively on the language side, but many
challenges might be best solved jointly (such as learning about the
new cereal appearance \emph{together} with the word to denote it).

\subsection{Environmental constraints}

% For agents to do something useful, they must live in an environment
% that is in some way ``natural'', such as the house-chores one I
% sketched above, not just designed for us entirely for purposes of
% simulation. I don't want to be too specific here, because I do not
% want to tie myself to the constraints of a specific use case, but I
% expect the following two aspects to be shared by virtually all
% interesting scenarios (because, if they do not hold, we are basically
% facing a problem that could be solved by traditional, hand-coded,
% logic-type AI).

Considering the use case above, the protolanguage should be
robust to the following environmental characteristics:

\begin{enumerate}
    \item Agent inputs are continuous and noisy: they might be (representations of) images, actions, sounds, etc. Different instances of the same object (or action, or property) will look at least slightly different, and boundaries between different categories are fuzzy.
    \item There is no close set of ``things'' that agents need to
      talk about: new objects, actions, properties can always appear.
\end{enumerate}

\subsection{Communication needs}

At a minimum, an agent might need to inform other
agents of the presence of something, as in:

\begin{exe}
	\ex A rat!
    \label{tiger}
\end{exe}

However, most useful messages will highlight a property of an object that is important for current purposes, as in:

\begin{exe}
	\ex \begin{xlist}
		\ex rat big \label{predication-rat-big}
    	\ex cat dead
	    \ex sister running \label{predication-running}
	    \ex child hungry
    	\ex apple eaten
	    \end{xlist}
    	\label{predication}
\end{exe}

I will call the structure exemplified in (\ref{predication}), in which a property is asserted of an object, a \emph{predication}. I am using ``property'' in a very broad sense. In particular, it can be a feature that natural languages would likely denote through an adjective, but it could also be an action or state typically denoted by a verb. Indeed, even example (\ref{tiger}) above could be seen as a concise way to express a predication akin to (\ref{tigerpred}):

\begin{exe}
	\ex rat near
    \label{tigerpred}
\end{exe}

Agents can go a long way with basic predication. In the next few sections I will discuss more complicated things that natural language can also do, which I think we can either \begin{inparaenum}[i)] 
\item ignore for the time being (Sec.~\ref{sec:ignore}), or
\item expect to be handled by context in the situated, goal-driven communication settings we envisage  (Sec.~\ref{sec:context})  , or
\item capture through simple predication and massive
  parataxis  (Sec.~\ref{sec:extreme-predication}).\footnote{In linguistics, \emph{parataxis} defines the tendency
    to juxtapose simple clauses, instead of subordinating them to construct
    complex sentences. Cf.~``I came, I saw, I conquered''
    vs.~``After coming, having seen, I decided to conquer''.}
\end{inparaenum}

\subsubsection{Language feats we can ignore for the time being}
\label{sec:ignore}

Being able to refer to abstract concepts is a huge part of what makes language so powerful. Yet, I conjecture that agents can be very useful without the need to think in abstract terms. Moreover, being able to talk about concrete things is probably a necessary stepping stone towards abstract language (\cite{Lakoff:Johnson:1999}).

We can moreover let the language be quite rough in its ways of reference, akin to the children's two-word stage. Let's forget for now about the nuances afforded by adverbial and clausal modification.

Finally, the protolanguage need not support the further expressive power complex verbal morphology and sentential scaffoldings would provide: for example, the ability to express counterfactuals, hypotheticals, belief degrees, etc.

\subsubsection{Context can provide a lot of side information}
\label{sec:context}

Primitive talking agents will have to deal with a relatively small
number of environments and tasks, where a relatively stable
extra-linguistic context can provide a lot of information. If our
home-bound agents are speaking of chicken, it's definitely the food
and not the animal. Consequently, for the time being, we probably do not need to
worry about (\emph{inter alia}):

\paragraph{Genericity.} Does ``rat big'' mean that rats are big in general, or that there is a big rat in the kitchen? Does ``apple eaten'' mean that apples are edible, or that someone is eating an apple? At least in a relatively restricted domain, there might be only one natural reading for each expression, or the intended meaning might be clear from context. Similarly, issues of definiteness (do we want that specific apple, or any apple will do?) can be left to extra-linguistic means of disambiguation.

\paragraph{Speech acts.} In (\ref{predication-rat-big}), are you asking me if the rat is big, or informing me that it is? When uttering (\ref{predication-running}), are you stating that your sister is running, or are you ordering her to? Again, context (who is the addressee, what are the current information needs, etc.) can take care of that.

\subsubsection{Pushing simple predication to the limit}
\label{sec:extreme-predication}

\paragraph{Expressing more complex concepts through massive parataxis.}

The agents can concatenate multiple predications to articulate more complex concepts, leaving the appropriate binding to context. Consider for example two-argument predicates, such as transitive verbs. As a primitive surrogate, our protolanguage could have separate predicates for the agent and theme roles, and closely juxtapose two predications to express the composite concept. So, ``a rat is eating the cat'' could be expressed by something like:

\begin{exe}
	\ex rat eating\ldots{} cat eaten
\end{exe}

Similarly, temporal or causal links could be expressed by parataxis aided by iconicity (that is, ordering predications in the way in which the corresponding events occurred, placing the cause before the effect, etc.):

\begin{exe}
	\ex \begin{xlist}
	\ex rat running\ldots{} cheese reached\ldots{} rat stopping \label{hunter-walks-stops}
	\ex rat hungry\ldots{} rat eating
	\end{xlist}\label{temporal-causal}
\end{exe}

Coreference can be left implicit or provided by sheer repetition of the same nouns, as in the examples in (\ref{temporal-causal}). Similar strategies will break down when a certain level of complexity is reached, but we can probably go a long way before we get there.

\paragraph{Personal pronouns, proper nouns,
  quantification.} A lot of things that
formal semanticists have very rightly argued cannot be treated as
standard nouns or adjectives \emph{could} be treated as standard nouns
or adjectives in our protolanguage (when the relevant meaning facets
cannot directly be left to context).\footnote{Formal semanticists with a heart condition
  might want to stop reading at this point.} The list
includes \begin{inparaenum}[i)]
\item personal pronouns, treated just like any other noun, except that their reference will be context-dependent (``me hungry'');
\item proper nouns, that, to the horror of philosophers of language, could be treated nouns that happen to denote one entity only (``Marco hungry''); and 
\item quantities, such as ``many'' and ``three'', that would be treated just like other property-denoting predicates (``rat many'').
\end{inparaenum} 

\section{Priorities for emergent language research}

To get us to the stage outlined in the previous section, we need the emergent language to have the following characteristics:

\begin{itemize}
\item use words\footnote{\emph{Morphemes} is probably a more accurate
    term here, but let's not get too technical.} to \emph{categorize} inputs lying on a continuous
  space into distinct classes (different \emph{objects}, such as
  \emph{rats}, \emph{actions}, such as \emph{running}, and
  \emph{properties}, such as \emph{being yellow});
\item seamlessly create new words when new concepts are encountered;
\item express combinations of arguments and predicates.
\end{itemize}

I will next discuss each of these priorities in turn, and conclude the
section with some remarks about the issue of making the protolanguage
human-understandable.

\subsection{Categorization}

You never step into the same river twice, but you still conceptualize
it as a single river; the way a kangaroo jumps is very different from
the way a grasshopper jumps; we call a broad range of animals
``dogs'', but it is useful to think of wolves and hyenas as
categorically different. Languages strike a good balance between
splitting the world into classes that are so detailed as to be of no
use (my keyboard at 3.30pm on February 14th 2020) and classes that are
so broad as to be again useless (``entity''). There is however little
ongoing language emergence research on how agents categorize their
perceptual input through language. The focus in the area has recently
shifted to toy environments where the input is either already
conveniently split into symbolically represented categories, or in any
case rather artificial, e.g., synthetic images of geometric shapes
(\citealp{Lazaridou:etal:2018} showed that already using synthetic
images instead of purely symbolic inputs has a big impact on the
emerging language).

\subsection{Word coinage}

Agents should be able to refer to new things that might appear on their horizon, such as a new type of cereal (or new ways to categorize existing things, like when you learned that a subset of the grains you already knew were cereals).

One way to do this is to provide the agents with a very large
vocabulary of primitive symbols (say, 20k symbols, a conservative
estimate of the lexical knowledge of an average English
speaker),\footnote{\url{https://bit.ly/2T7CVOP}} so that they can recruit hitherto unused symbols to refer
to new concepts on demand.

A more human-like (and probably more promising) approach is to lead agents to
exploit \emph{duality of patterning} (\cite{Martinet:1965}), that is,
treating primitive symbols as meaningless units to be combined into
meaningful words. Agents are provided with a small set of symbols,
more akin to a phoneme inventory or alphabet, so that, from the very
start, they must make up words by concatenating symbols from this
alphabet. Even limiting alphabet size and maximum utterance length,
this immediately provides the potential to coin a very large number of
words. % \footnote{Given maximum length $L$ and alphabet size $S$, the
%   number of possible words is $\sum_{x=1}^{x=L} S^x$, so small values
%   such as $L=4$ and $S=10$ already suffice to generate more than 10k
%   distinct words.} % Since the agents would soon exhaust the number of
% % concepts they can refer to with single symbols, they would have, from
% very early on, to engage in new-word coinage.
% %  With this approach, agents would be using all
% symbols from the very start, whereas under the one above many symbols
% would lie unused until needed. We know that initializing embeddings
% for unused symbols is something that comes hard to neural networks
% \citep{Herbelot:Baroni:2017}, so it's good if we can avoid
% it. Also, the alphabet-based approach might lead to the emergence of interesting phenomena, such as a proto-morphology, if agents
% start using similar strings for similar concepts.

New word coinage is not systematically studied in the current language
emergence literature. What is extensively studied instead is the case
where new referents are composite, and the question is whether the
emergent language can harness this composite structure to quickly
generalize (aka \emph{compositionality}, see, e.g.,
\cite{Kottur:etal:2017,Lazaridou:etal:2018,Mordatch:Abbeel:2018,Andreas:2019,Guo:etal:2019,Resnick:etal:2020}). This
is of course an important question, but simple-word coinage should
come first. From an evolutionary perspective, it only makes sense for
a language to start composing words in order to denote complex
meanings after the language has coined enough primitive words to
justify combinatorial strategies. From a methodological perspective,
if we do not understand the means a language uses to refer to new
primitive concepts, we will be doomed when trying to analyze how it
refers to composite ones. Without an understanding of basic word
coinage, it will be hard to tell whether, when an agent utters
``baama'' in response to a red triangle, it is using ``baa'' to refer
to redness and ``ma'' to refer to triangles, or whether it is treating
the whole as new primitive concept that it is labeling with the string
``baama''.

\subsection{Argument-predicate combinations}

Agents able to communicate about an open set of primitive categories
by adding new words to their repertoire would already be quite a
feat. The expressiveness of the language would however enormously
increase if object-denoting and action-state-or-property-denoting
words could be combined in simple predication structures. This is
close to what is currently studied under the rubric of
compositionality in emergent language (see above).  However, there are
some aspects of compositionality that I would like to prioritize,
based on the desiderata for the protolanguage I have outlined.

Most importantly, if we want to develop a language that can eventually
be deployed in the real world, the agents should be able to detect the properties to be denoted by predicates in perceptually
realistic inputs, rather than in abstract symbolic
representations.  It is one thing to learn a \emph{red} predicate from
input objects represented by attribute-value sets such as $\{$\emph{color:RED},
  \emph{shape:SQUARE}$\}$, and another to discover that red hair and red
tomatoes, as depicted in natural images, have something in common that
can be captured by a shared \emph{red} predicate.  Decomposing a dynamic
scene into an agent or patient and the action or event they are
involved in is even more challenging (and could perhaps be left to a
later stage of research).

% I conjectured above that utterances expressing a single
% one-argument predication can go a long way. As I discussed,
% the first step towards referring to more complex situations can be to encourage agents to produce multiple single-predication utterances (``rat
% eating \ldots cat eaten''). In other words, the language might feature a single binary composition operation only, but agents can resort to this structure multiple times in the same paratactically complex act of reference. Realistically, though, we have enough on our plate already with a single simple predication, to defer the case of multiple predications to later stages of investigation. Needless to say, more complex composition phenomena such as subordination and recursion are best left to future work.

Scaling up, I described above a way to refer to more complex
situations by paratactic concatenation of unary predications (``rat
eating \ldots cat eaten''). I think however that we already have
enough on our plate dealing with the single predication case. Needless
to say, more ambitious forms of composition, such as clausal
subordination and recursion, are best left to future work.

\subsection{Humans in the protolanguage loop}
% from my use case.
In the framework of \emph{emergent} communication, we do not directly
control the actual form agent language will take. Consequently, the
linguistic examples I presented above are just human-friendly
illustrations of the structures I'd like to see in the emergent
protolanguage. The way the same communicative needs are satisfied by
the protolanguage could be very different.

Since one of our goals is human-machine interaction, if the
protolanguage drifts too far from anything human-like (consider a
language that achieves composition by symbol inter-leaving: ``rceadr''
denoting a red car), we'll need to bring it back to Terran grounds
somehow. We can take here inspiration from current research on
limiting emergent language drift, either by forcing the agents to
directly mimic natural language
(\cite{Havrylov:Titov:2017,Lazaridou:etal:2017,Lee:etal:2018b}), or by
imposing human-like bottlenecks, for example, on agent memory and
channel capacity (\cite{Kottur:etal:2017,Resnick:etal:2020}). More generally, we can find inspiration in the growing area of model interpretability and explainability (e.g., \cite{Ribeiro:etal:2016}).

Still, just as we have no big trouble interacting with children, or to
quickly pick up fundamental words and constructions in a foreign
language, once machines are endowed with a minimally sensible
protolanguage, I expect humans would be willing and able to make an
effort to understand what they mean.

\section{How do we get there?}

Let's recall, again, that we are considering the emergent language
setup, where agents develop a communication protocol in order to solve
a task together, without any supervision on the protocol itself. I am
\emph{not} proposing to force the properties I outlined above into the
emergent language by manual coding or \emph{ad-hoc} training. Decades
of failed attempts in ML/AI suggest that manual language coding is
invariably a bad idea, as shown by the current dominance of end-to-end
deep networks over systems relying on explicit linguistic structures
in virtually all domains of natural language processing (see, e.g.,
the state of the art in tasks such as machine translation, and machine
comprehension; \cite{Cui:etal:2017,Edunov:etal:2018}). At the same
time, coming up with ``protolanguage'' data to train agents is a
contrived and difficult enterprise (the whole idea of language
emergence arose in the deep NLP community because it is not clear how
to train an interactive talking agent in a supervised way!). The focus
should be instead on designing environments and tasks/games in which
the agents are naturally encouraged to develop the properties of
interest.\footnote{Coming up with proper evaluation methods is equally
  important.  The latter should not only quantify task success, but
  also the properties of the emergent language we find desirable,
  which is a huge challenge in and by itself
  (\cite{Bouchacourt:Baroni:2018,Lazaridou:etal:2018,Lowe:etal:2019}).}

Coming back to my application scenario above, a simplified laboratory
setup could involve multiple agents that have to jointly maintain a
set of household goods in stock. Minimally, there are two agents. A
\emph{pantry} agent keeps track of the current supply level, that is
constantly in flux as some items are consumed, others go bad, etc.. A
\emph{shopping} agent would have to provide missing goods, possibly
with budgetary restrictions.

In a first iteration, the game could mix realistic components and
opportunistic choices dictated by practical considerations. For
example, the distribution of natural images of household goods could
be taken from a database such as
LVIS,\footnote{\url{https://www.lvisdataset.org/}} where objects have
a realistic long-tail distribution. To encourage predication,
attribute classifiers could automatically tag the images with
properties such as color (\cite{Fahradi:etal:2009}), and the stock
could be built to contain many objects only differing in some
attribute value (e.g., candies of different colors). On the other
hand, changes in stock would be guided by random processes; change of
state (e.g., rotting) could be simulated by techniques to deform
shapes in pictures, and so on.
This scenario should meet my desiderata: realistic perceptual input,
new objects steadily popping up, at least some degree of predication
needed (``tomatoes green'', ``potatoes out'', ``mozzarella
rotten''). However, I recognize that my sketch is rather hand-wavy. I
hope these notes will inspire other researchers to brainstorm together
about better ideas on how to get started.

\section*{Acknowledgments}

I thank Gemma Boleda, Diane Bouchacourt, Thomas Brochhagen and
Jean-R\'emi King for feedback on a draft of this roadmap. I also thank
Rahma Chaabouni, Roberto Dess\`i, Emmanuel Dupoux, Eugene Kharitonov, Angeliki
Lazaridou and Yann LeCun for various forms of helpful brainstorming.

\bibliographystyle{plainnat}
\bibliography{marco}

\begin{thebibliography}{31}
\providecommand{\natexlab}[1]{#1}
\providecommand{\url}[1]{\texttt{#1}}
\expandafter\ifx\csname urlstyle\endcsname\relax
  \providecommand{\doi}[1]{doi: #1}\else
  \providecommand{\doi}{doi: \begingroup \urlstyle{rm}\Url}\fi

\bibitem[Andreas(2019)]{Andreas:2019}
Jacob Andreas.
\newblock Measuring compositionality in representation learning.
\newblock In \emph{Proceedings of ICLR}, New Orleans, LA, 2019.
\newblock Published online:
  \url{https://openreview.net/group?id=ICLR.cc/2019/conference}.

\bibitem[Berk and {Lillo-Martin}(2012)]{Berk:LilloMartin:2012}
Stephanie Berk and Diane {Lillo-Martin}.
\newblock The two-word stage: Motivated by linguistic or cognitive constraints?
\newblock \emph{Cognitive Psychology}, 65\penalty0 (1):\penalty0 118--140,
  2012.

\bibitem[Bickerton(2014)]{Bickerton:2014}
Derek Bickerton.
\newblock \emph{More than Nature Needs: Language, Mind, and Evolution}.
\newblock Harvard University Press, Cambridge, MA, 2014.

\bibitem[Bloom(1970)]{Bloom:1970}
Lois Bloom.
\newblock \emph{Language Development: Form and Function in Emerging Grammars}.
\newblock MIT Press, Cambridge, MA, 1970.

\bibitem[Bouchacourt and Baroni(2018)]{Bouchacourt:Baroni:2018}
Diane Bouchacourt and Marco Baroni.
\newblock How agents see things: On visual representations in an emergent
  language game.
\newblock In \emph{Proceedings of EMNLP}, pages 981--985, Brussels, Belgium,
  2018.

\bibitem[Brentari and {Goldin-Meadow}(2017)]{Brentari:GoldinMeadow:2017}
Diane Brentari and Susan {Goldin-Meadow}.
\newblock Language emergence.
\newblock \emph{Annual Review of Linguistics}, 3:\penalty0 617--645, 2017.

\bibitem[Brochhagen(2018)]{Brochhagen:2018}
Thomas Brochhagen.
\newblock \emph{Signaling under uncertainty}.
\newblock Ph.{D} dissertation, University of Amsterdam, 2018.

\bibitem[Cao et~al.(2018)Cao, Lazaridou, Lanctot, Leibo, Tuyls, and
  Clark]{Cao:etal:2018}
Kris Cao, Angeliki Lazaridou, Marc Lanctot, Joel Leibo, Karl Tuyls, and Stephen
  Clark.
\newblock Emergent communication through negotiation.
\newblock In \emph{Proceedings of ICLR Conference Track}, Vancouver, Canada,
  2018.
\newblock Published online:
  \url{https://openreview.net/group?id=ICLR.cc/2018/Conference}.

\bibitem[Chaabouni et~al.(2019)Chaabouni, Kharitonov, Dupoux, and
  Baroni]{Chaabouni:etal:2019}
Rahma Chaabouni, Eugene Kharitonov, Emmanuel Dupoux, and Marco Baroni.
\newblock Anti-efficient encoding in emergent communication.
\newblock In \emph{Proceedings of NeurIPS}, Vancouver, Canada, 2019.
\newblock Published online:
  \url{https://papers.nips.cc/book/advances-in-neural-information-processing-systems-32-2019}.

\bibitem[Cui et~al.(2017)Cui, Chen, Wei, Wang, Liu, and Hu]{Cui:etal:2017}
Yiming Cui, Zhipeng Chen, Si~Wei, Shijin Wang, Ting Liu, and Guoping Hu.
\newblock Attention-over-attention neural networks for reading comprehension.
\newblock In \emph{Proceedings of ACL}, pages 593--602, Vancouver, Canada,
  2017.

\bibitem[Edunov et~al.(2018)Edunov, Ott, Auli, and Grangier]{Edunov:etal:2018}
Sergey Edunov, Myle Ott, Michael Auli, and David Grangier.
\newblock Understanding back-translation at scale.
\newblock In \emph{Proceedings of EMNLP}, pages 489--500, Brussels, Belgium,
  2018.

\bibitem[Evtimova et~al.(2018)Evtimova, Drozdov, Kiela, and
  Cho]{Evtimova:etal:2018}
Katrina Evtimova, Andrew Drozdov, Douwe Kiela, and Kyunghyun Cho.
\newblock Emergent communication in a multi-modal, multi-step referential game.
\newblock In \emph{Proceedings of ICLR Conference Track}, Vancouver, Canada,
  2018.
\newblock Published online:
  \url{https://openreview.net/group?id=ICLR.cc/2018/Conference}.

\bibitem[Farhadi et~al.(2009)Farhadi, Endres, Hoiem, and
  Forsyth]{Fahradi:etal:2009}
Ali Farhadi, Ian Endres, Derek Hoiem, and David Forsyth.
\newblock Describing objects by their attributes.
\newblock In \emph{Proceedings of CVPR}, pages 1778--1785, Miami Beach, FL,
  2009.

\bibitem[Foerster et~al.(2016)Foerster, Assael, de~Freitas, and
  Whiteson]{Foerster:etal:2016}
Jakob Foerster, Ioannis~Alexandros Assael, Nando de~Freitas, and Shimon
  Whiteson.
\newblock Learning to communicate with deep multi-agent reinforcement learning.
\newblock In \emph{Proceedings of NIPS}, pages 2137--2145, Barcelona, Spain,
  2016.

\bibitem[Guo et~al.(2019)Guo, Ren, Havrylov, Frank, Titov, and
  Smith]{Guo:etal:2019}
Shangmin Guo, Yi~Ren, Serhii Havrylov, Stella Frank, Ivan Titov, and Kenny
  Smith.
\newblock The emergence of compositional languages for numeric concepts through
  iterated learning in neural agents.
\newblock \url{https://arxiv.org/abs/1910.05291}, 2019.

\bibitem[Havrylov and Titov(2017)]{Havrylov:Titov:2017}
Serhii Havrylov and Ivan Titov.
\newblock Emergence of language with multi-agent games: Learning to communicate
  with sequences of symbols.
\newblock In \emph{Proceedings of NIPS}, pages 2149--2159, Long Beach, CA,
  2017.

\bibitem[Hurford(2014)]{Hurford:2014}
James Hurford.
\newblock \emph{The Origins of Language}.
\newblock Oxford University Press, Oxford, UK, 2014.

\bibitem[Jackendoff and Wittenberg(2014)]{Jackendoff:Wittenberg:2014}
Ray Jackendoff and Eva Wittenberg.
\newblock What you can say without syntax: A hierarchy of grammatical
  complexity.
\newblock In Frederick Newmeyer and Laurel Preston, editors, \emph{Measuring
  Grammatical Complexity}, pages 65--82. Oxford University Press, Oxford, UK,
  2014.

\bibitem[Kottur et~al.(2017)Kottur, Moura, Lee, and Batra]{Kottur:etal:2017}
Satwik Kottur, Jos{\'e} Moura, Stefan Lee, and Dhruv Batra.
\newblock Natural language does not emerge `naturally' in multi-agent dialog.
\newblock In \emph{Proceedings of EMNLP}, pages 2962--2967, Copenhagen,
  Denmark, 2017.

\bibitem[Lakoff and Johnson(1999)]{Lakoff:Johnson:1999}
George Lakoff and Mark Johnson.
\newblock \emph{Philosophy in the Flesh: The Embodied Mind and Its Challenge to
  Western Thought}.
\newblock Basic Books, New York, 1999.

\bibitem[Lazaridou and Baroni(2020)]{Lazaridou:Baroni:2020}
Angeliki Lazaridou and Marco Baroni.
\newblock Emergent multi-agent communication in the deep learning era, 2020.
\newblock {U}nder review.

\bibitem[Lazaridou et~al.(2017)Lazaridou, Peysakhovich, and
  Baroni]{Lazaridou:etal:2017}
Angeliki Lazaridou, Alexander Peysakhovich, and Marco Baroni.
\newblock Multi-agent cooperation and the emergence of (natural) language.
\newblock In \emph{Proceedings of ICLR Conference Track}, Toulon, France, 2017.
\newblock Published online:
  \url{https://openreview.net/group?id=ICLR.cc/2017/conference}.

\bibitem[Lazaridou et~al.(2018)Lazaridou, Hermann, Tuyls, and
  Clark]{Lazaridou:etal:2018}
Angeliki Lazaridou, {Karl Moritz} Hermann, Karl Tuyls, and Stephen Clark.
\newblock Emergence of linguistic communication from referential games with
  symbolic and pixel input.
\newblock In \emph{Proceedings of ICLR Conference Track}, Vancouver, Canada,
  2018.
\newblock Published online:
  \url{https://openreview.net/group?id=ICLR.cc/2018/Conference}.

\bibitem[Lee et~al.(2018)Lee, Cho, Weston, and Kiela]{Lee:etal:2018b}
Jason Lee, Kyunghyun Cho, Jason Weston, and Douwe Kiela.
\newblock Emergent translation in multi-agent communication.
\newblock In \emph{Proceedings of ICLR Conference Track}, Vancouver, Canada,
  2018.
\newblock Published online:
  \url{https://openreview.net/group?id=ICLR.cc/2018/Conference}.

\bibitem[Li and Bowling(2019)]{Li:Bowling:2019}
Fushan Li and Michael Bowling.
\newblock Ease-of-teaching and language structure from emergent communication.
\newblock In \emph{Proceedings of NeurIPS}, Vancouver, Canada, 2019.
\newblock Published online:
  \url{https://papers.nips.cc/book/advances-in-neural-information-processing-systems-32-2019}.

\bibitem[Lowe et~al.(2019)Lowe, Foerster, Boureau, Pineau, and
  Dauphin]{Lowe:etal:2019}
Ryan Lowe, Jakob Foerster, {Y-Lan} Boureau, Joelle Pineau, and Yann Dauphin.
\newblock On the pitfalls of measuring emergent communication.
\newblock In \emph{Proceedings of AAMAS}, pages 693--701, Montreal, Canada,
  2019.

\bibitem[Martinet(1965)]{Martinet:1965}
Andr\'e Martinet.
\newblock \emph{La Linguistique Synchronique: \'{E}tudes et Recherches}.
\newblock Presses Universitaires de France, Paris, France, 1965.

\bibitem[Mordatch and Abbeel(2018)]{Mordatch:Abbeel:2018}
Igor Mordatch and Pieter Abbeel.
\newblock Emergence of grounded compositional language in multi-agent
  populations.
\newblock In \emph{Proceedings of AAAI}, pages 1495--1502, New Orleans, LA,
  2018.

\bibitem[Radford et~al.(2019)Radford, Wu, Child, Luan, Amodei, and
  Sutskever]{Radford:etal:2019}
Alec Radford, Jeffrey Wu, Rewon Child, David Luan, Dario Amodei, and Ilya
  Sutskever.
\newblock Language models are unsupervised multitask learners.
\newblock
  \url{https://d4mucfpksywv.cloudfront.net/better-language-models/language-models.pdf},
  2019.

\bibitem[Resnick et~al.(2020)Resnick, Gupta, Foerster, Dai, and
  Cho]{Resnick:etal:2020}
Cinjon Resnick, Abhinav Gupta, Jakob Foerster, Andrew Dai, and Kyunghyun Cho.
\newblock Capacity, bandwidth, and compositionality in emergent language
  learning.
\newblock In \emph{Proceedings of AAMAS}, Auckland, New Zealand, 2020.
\newblock {I}n press.

\bibitem[Ribeiro et~al.(2016)Ribeiro, Singh, and Guestrin]{Ribeiro:etal:2016}
Marco Ribeiro, Sameer Singh, and Carlos Guestrin.
\newblock ``{Why} should {I} trust you?'': Explaining the predictions of any
  classifier.
\newblock In \emph{Proceedings of KDD}, pages 1135--1144, San Francisco, CA,
  2016.

\end{thebibliography}
\end{document}